\documentclass[conference,letter,9.96pt]{IEEEtran}
\usepackage{cite}
\usepackage{amsmath,amssymb,amsfonts}
\usepackage{algorithmic}
\usepackage{graphicx}
\usepackage{textcomp}
\usepackage{xcolor}
\usepackage{stfloats}

\usepackage[left=0.75in,right=0.76in,top=0.820in, bottom=1.287in]{geometry}

\usepackage{hhline}
\usepackage{makecell}
\usepackage{booktabs}
\usepackage[flushleft]{threeparttable}

\DeclareMathOperator{\sech}{sech}

\def\BibTeX{{\rm B\kern-.05em{\sc i\kern-.025em b}\kern-.08em
    T\kern-.1667em\lower.7ex\hbox{E}\kern-.125emX}}
\begin{document}

\title{On the Computational Complexities of
Complex-valued Neural Networks}

\author{\IEEEauthorblockN{Kayol S. Mayer, Jonathan A. Soares, Ariadne A. Cruz, and Dalton S. Arantes}
\IEEEauthorblockA{\textit{Department of Communications, School of Electrical and Computer Engineering} \\
\textit{Universidade Estadual de Campinas -- UNICAMP}, Campinas, SP, Brazil\\
kayol@unicamp.br, j229966@dac.unicamp.br, a038481@dac.unicamp.br, dalton@unicamp.br}
}

\maketitle


\begin{abstract}
Complex-valued neural networks (CVNNs) are nonlinear filters used in the digital signal processing of complex-domain data. Compared with real-valued neural networks~(RVNNs), CVNNs can directly handle complex-valued input and output signals due to their complex domain parameters and activation functions. With the trend toward low-power systems, computational complexity analysis has become essential for measuring an algorithm’s power consumption. Therefore, this paper presents both the quantitative and asymptotic computational complexities of CVNNs. This is a crucial tool in deciding which algorithm to implement. The mathematical operations are described in terms of the number of real-valued multiplications, as these are the most demanding operations. To determine which CVNN can be implemented in a low-power system, quantitative computational complexities can be used to accurately estimate the number of floating-point operations. We have also investigated the computational complexities of CVNNs discussed in some studies presented in the literature.
\end{abstract}



\section{Introduction}
Since the first steps of artificial neural models, a significant number of artificial neural network~(ANN) architectures and learning methods have been proposed~\cite{Albawi2017,Guo2017,Carlini2017}. Interestingly, among these artificial neural networks, scarce attention is paid to the class of complex-valued neural networks~(CVNNs)~\cite{mayer2022b}. Unlike real-valued neural networks (RVNNs), CVNNs are capable of directly handling complex inputs and outputs [5]. As a result, CVNNs should be the natural choice for processing complex-valued signals, and they should also be explored for real-valued applications. Take for instance the XOR problem, derived from the two-dimensional “AND/OR” theorem. A single real-valued perceptron is unable to learn the XOR function. To solve the XOR problem, a three-layer RVNN is necessary at the very least. However, Minsky and Papert’s limitation can be circumvented using only a single complex-valued neuron ~\cite{Nitta2003}. 
Yet, the use of a single complex-valued neuron is not the only motivation; with CVNN architectures, it's possible to enhance the functionality of neural networks, improve their performance, and reduce training time compared to RVNNs ~\cite{Hirose2012b,Zhang2022}. Furthermore, it was recently proven by Voigtlaender~\cite{Voigtlaender2023} that CVNNs also adhere to the universal approximation theorem. 


For real-time systems, CVNNs have recently been implemented in photonic integrated circuits as an optical neural chip that obtained faster convergence and higher accuracy compared with RVNNs~\cite{Zhang2021}. Not only in optical neural chips, CVNNs can also be efficiently implemented in graphics processing units~(GPUs) and tensor processing units~(TPUs) with matrix structures and field programmable gate arrays~(FPGAs) with systolic arrays~\cite{Yang2022}. Additionally, with the development of adaptive computing platforms, such as PYNQ from Xilinx~\cite{pynq}, CVNNs can be easily implemented in hardware using open-source Python libraries~(e.g., RosenPy, developed by Cruz et al.~\cite{Cruz2022}).

In many current applications, the most demanding algorithms are usually centralized in base stations with significant computational power. However, new technologies claim for desegregation, such as the Internet of Things~(IoT), smart homes, and Industry 4.0, where a significant number of intelligent sensors are necessary~\cite{Pech2021}. Then, computational complexity analysis is crucial to choose the best approach for low-power systems. 

For digital communication systems, CVNNs have also presented promising results for telecommunications, such as channel estimation and equalization, beamforming, detection, and decoding~\cite{Liu2019, Enriconi2020, Mayer2020a, Soares2021, Xu2022, Mayer2022, Chu2022,Soares2023}. Liu et al.~\cite{Liu2019} proposed a CVNN based on extreme learning machines for channel estimation and equalization for OFDM systems. Enriconi et al.~\cite{Enriconi2020} demonstrated the beamforming tracking performance of a shallow phase transmittance radial basis function~(PT-RBF) neural network under a dynamic military channel. Mayer et al.~\cite{Mayer2020a} employed a modified PT-RBF for transmitting beamforming, including the array currents into the CVNN architecture. Soares et al.~\cite{Soares2021} implemented a joint channel estimation and decoding for massive-MIMO communications using a shallow PT-RBF. Xu et al.~\cite{Xu2022} applied deep convolutional CVNNs for raw IQ signal recognition, achieving improved accuracy with lower computation complexity compared with RVNNs. Mayer et al.~\cite{Mayer2022} compared some CVNN architectures for receiver beamforming operating with multiple users and interferences. Chu et al.~\cite{Chu2022} proposed a channel estimation technique using a CVNN for optical systems operating with filter bank multicarrier with offset quadrature amplitude modulation~(FBMC/OQAM). Soares et al.~\cite{Soares2023} proposed two inference learning approaches for channel estimation and decoding with CVNNs under highly dynamic channels.

In the literature, some CVNN computational complexities are addressed depending on the system implementation. In \cite{Mayer2019a}, the computational complexity of a shallow PT-RBF is presented in terms of a concurrent equalizer and a fuzzy controller. In~\cite{Soares2021}, the computational complexity of a shallow PT-RBF is described as a function of the MIMO communication architecture. To the best of our knowledge, there is no work comparing the computational complexities of CVNNs, such as complex-valued feedforward NN~(CVFNN)~\cite{Dong2021}, split-complex feedforward NN~(SCFNN)~\cite{Scardapane2020}, multilayer feedforward NN based on multi-valued neurons~(MLMVN)~\cite{Aizenberg2016}, complex-valued radial basis function~(C-RBF)~\cite{Enriconi2020}, fully complex-valued radial basis function~(FC-RBF)~\cite{Savitha2012}, and PT-RBF~\cite{Mayer2022}.

This paper is an extension of Kayol S. Mayer's Ph.D. Thesis~\cite{mayer2022b}, developed at the School of Electrical and Computer Engineering, Universidade Estadual de Campinas, in the area of Telecommunications and Telematics. In this context, this paper presents the quantitative and asymptotic computational complexities of the mentioned CVNNs in a comprehensive way, regardless of any specific application.

The remainder of this paper is organized as follows. Section~\ref{sec:cvnns} presents a brief discussion on CVNNs. Section~\ref{sec:CompComplex} describes the quantitative and asymptotic computational complexities of CVNNs. In Section~\ref{sec:applcases}, we discuss the computational complexities of CVNNs proposed in the literature. Lastly, Section~\ref{sec:conc} concludes the paper.

\section{Complex-valued Neural Networks}
\label{sec:cvnns}

One of the most studied CVNNs in the literature is the complex-valued feedforward neural network, a multilayer perceptron without feedback among layers in the forward step, adapted to directly process data in the complex domain \cite{Dong2021}. CVFNNs can operate with fully-complex transcendental activation functions that satisfy the Cauchy-Riemann equations with relaxed conditions, such as circular, inverse circular, hyperbolic, and inverse hyperbolic functions. Also, an important and particular case of CVFNNs is the SCFNN, in which real and imaginary components are processed separately by holomorphic functions (i.e., analytic functions) in $\mathbb{R}$ \cite{Scardapane2020}. With similar architecture, but utilizing phase mappings onto unit circles as activation functions, the MLMVN is another relevant CVNN. In the MLMVN, the backpropagation algorithm is performed only using the multi-valued neurons error since no derivative is necessary because it is impossible to move in incorrect directions \cite{Aizenberg2016}. 

Based on a different CVNN architecture, the C-RBF neural network can also operate with complex numbers \cite{Enriconi2020}. Due to the C-RBF phase vanishing into the Euclidean norm of Gaussian neurons, Savitha et al.~\cite{Savitha2009} proposed the FC-RBF neural network, where $\sech(\cdot)$ activation functions map $\mathbb{C}^{N}\mapsto\mathbb{C}$ with Gaussian-like characteristics. Considering split-complex Gaussian neurons to circumvent any phase issue~\cite{Mayer2019a}, Loss et al.~\cite{Loss2007} proposed the shallow and multiple-input single-output (MISO) PT-RBF. Recently, the PT-RBF has been extended to multiple outputs~\cite{Soares2021} and multiple layers~\cite{Mayer2022}.

In communication systems, the choice of architecture for complex-valued neural networks (CVNNs) can significantly affect performance. Notably, as detailed in Mayer~\cite{mayer2022b}, our research has shown that RBF-based architectures consistently outperform other CVNN architectures, especially in communication-related tasks such as channel equalization, beamforming, channel estimation, and decoding. This superior performance can be attributed to the inherent characteristics of RBF-based neural networks that make them well-suited for handling additive white Gaussian noise (AWGN), a common feature in communication systems. This phenomenon can be understood by the similarity between the activation functions of RBF-based CVNNs and the distribution function of AWGN noise --- a distinction not found in other CVNNs like CVFNN, SCFNN, and MLMVN. This effect becomes more pronounced in challenging scenarios with lower signal-to-noise ratios~(SNRs). However, this does not apply to FC-RBF, which becomes unstable in noisy situations. Further insights regarding performance and parameter estimation are available in~\cite{mayer2022b, Mayer2022, Soares2023}.


\section{Computational Complexities}
\label{sec:CompComplex}

\subsection{Quantitative computational complexities}

In order to estimate the computational complexity of an algorithm, one of the more straightforward and effective strategies is the mathematical operations analysis. Based on the CVFNN, SCFNN, MLMVN, C-RBF, FC-RBF, and PT-RBF architectures proposed in the literature, the mathematical operations are summarized into additions, multiplications, and activation functions. Although activation functions encompass a set of nonlinear functions that seem burdensome at first glance, they are not considered in our analysis since lookup tables can efficiently implement them \cite{Zamanlooy2014}. For all CVNN architectures, the number of additions and multiplications are similar; thus, since the latter is much more demanding, additions are also not taken into account. For a complimentary analysis of the number of additions and activation functions of CVNNs, see~\cite{mayer2022b}.

The computational complexities of CVFNN, SCFNN, MLMVN, C-RBF, FC-RBF, and PT-RBF are assessed based on the number of inputs $P$, outputs $R$, layers $L$, and complex-valued artificial neurons per hidden layer $N$. Therefore, setting the number of inputs, outputs, and layers with neurons, we obtain the CVNNs computational complexities, depicted in Tables~\ref{tab:shallow_cvnn_comp_complexity} and \ref{tab:deep_cvnn_comp_complexity} for shallow and deep CVNNs, respectively. In Table~\ref{tab:deep_cvnn_comp_complexity}~, each CVNN layer is composed of $I^{\{l\}}$ complex-valued neurons for $l \in [1,\,2,\,\cdots,\,L-1]$, except for the input layer where $I^{\{l=0\}}=P$ and the output layer where $I^{\{L\}}=R$. Furthermore, for the deep PT-RBF, the number of bottleneck outputs is $O^{\{l\}}$.
It is important to notice that the C-RBF and FC-RBF are not taken into account in Table~\ref{tab:deep_cvnn_comp_complexity} because they are only proposed for shallow architectures.

\begin{table}[htbp]
\centering
\renewcommand{\arraystretch}{1.3}
\begin{threeparttable}
\caption{Shallow CVNN computational complexities.}
\label{tab:shallow_cvnn_comp_complexity}
\begin{tabular}{l c c}
\hline
\textbf{CVNN}& \textbf{Training} & \textbf{Inference}\\
\hline
CVFNN  & $N(8P+12R+8)+8R$   & $4N(P+R)$\\
SCFNN  & $N(8P+12R+8)+6R$   & $4N(P+R)$\\
MLMVN  & $N(8P+12R+16)+12R$ & $4N(P+R+1)+4R$\\
C-RBF  & $N(4P+6R+5)+4R$    & $N(2P+2R+1)$\\
FC-RBF & $N(12P+12R+12)+4R$ & $4N(P+R)$\\
PT-RBF & $N(4P+12R+12)+4R$  & $2N(P+2R+1)$\\ 
\hline
\end{tabular}
\end{threeparttable}
\end{table}

\begin{table*}[!b]
\centering
\renewcommand{\arraystretch}{1.3}
\begin{threeparttable}
\caption{Deep CVNN computational complexities.}
\label{tab:deep_cvnn_comp_complexity}
\begin{tabular}{l c c}
\hline
\textbf{CVNN}& \textbf{Training} & \textbf{Inference}\\
\hline
CVFNN  & 
$4\sum\limits_{l=1}^{L-1} I^{\{l\}}\left ( 2I^{\{l-1\}}+I^{\{l+1\}}+2\right )+8I^{\{L\}}\left (I^{\{L-1\}}+1\right )$
 & $\begin{aligned}[c] 
4\sum\limits_{l=1}^L I^{\{l\}}I^{\{l-1\}}
\end{aligned}$\\[0.3cm]
SCFNN  & $4\sum\limits_{l=1}^{L-1} I^{\{l\}}\left ( 2I^{\{l-1\}}+I^{\{l+1\}}+2\right )
+2I^{\{L\}}\left (4I^{\{L-1\}}+3\right )$ & $\begin{aligned}[c] 
4\sum\limits_{l=1}^L I^{\{l\}}I^{\{l-1\}}
\end{aligned}$\\[0.3cm]
MLMVN  & $4\sum\limits_{l=1}^{L-1} I^{\{l\}}\left ( 2I^{\{l-1\}}+I^{\{l+1\}}+4\right )+4I^{\{L\}}\left (2I^{\{L-1\}}+3\right )$ & $\begin{aligned}[c] 
4\sum\limits_{l=1}^L I^{\{l\}}\left(I^{\{l-1\}}+1\right)
\end{aligned}$\\[0.3cm]
PT-RBF & $4\sum\limits_{l=1}^{L} I^{\{l\}}\left ( O^{\{l-1\}}+3O^{\{l\}}+3\right )+ 4\sum\limits_{l=1}^{L-1}O^{\{l\}}\left(I^{\{l+1\}}+1\right)+4O^{\{L\}}$  & $\begin{aligned}[c] 
2\sum\limits_{l=1}^L I^{\{l\}}\left(O^{\{l-1\}}+2O^{\{l\}}+1\right)
\end{aligned}$\\ 
\hline
\end{tabular}
\end{threeparttable}
\end{table*}

\begin{table*}[!b]
\centering
\renewcommand{\arraystretch}{1.3}
\begin{threeparttable}
\caption{CVNN asymptotic computational complexities.}%
\label{tab:asymptotic_cvnn_comp_complexity}
\begin{tabular}{l c c | c c}
\hline
 & \multicolumn{2}{c}{\textbf{Shallow}} & \multicolumn{2}{c}{\textbf{Deep}}\\[0.05cm]
\hhline{~----}
\textbf{CVNN}& $\boldsymbol{P=R\ll N}$ & $\boldsymbol{P=R\approx N}$ & $\boldsymbol{P=R=N \gg L}$ & $\boldsymbol{P=R=N \approx L}$\\
\hline
CVFNN  & $\mathcal{O}\left(N\right)$ & $\mathcal{O}\left(N^2\right)$ & $\mathcal{O}\left(N^2\right)$ & $\mathcal{O}\left(N^3\right)$\\[0.3cm]
SCFNN  & $\mathcal{O}\left(N\right)$ & $\mathcal{O}\left(N^2\right)$ & $\mathcal{O}\left(N^2\right)$ & $\mathcal{O}\left(N^3\right)$\\[0.3cm]
MLMVN  & $\mathcal{O}\left(N\right)$ & $\mathcal{O}\left(N^2\right)$ & $\mathcal{O}\left(N^2\right)$ & $\mathcal{O}\left(N^3\right)$\\[0.3cm]
C-RBF  & $\mathcal{O}\left(N\right)$ & $\mathcal{O}\left(N^2\right)$ & $\mathbf{-}$ & $\mathbf{-}$\\[0.3cm]
FC-RBF & $\mathcal{O}\left(N\right)$ & $\mathcal{O}\left(N^2\right)$ & $\mathbf{-}$ & $\mathbf{-}$\\[0.3cm]
PT-RBF & $\mathcal{O}\left(N\right)$ & $\mathcal{O}\left(N^2\right)$ & $\mathcal{O}\left(N^2\right)$ & $\mathcal{O}\left(N^3\right)$\\ 
\hline
\end{tabular}
\begin{tablenotes}
\item $\mathbf{-}$ not applicable.
\end{tablenotes}
\end{threeparttable}
\end{table*}

\subsection{Asymptotic computational complexities} \label{sec:AsympComplexi}

We assume that the neural networks have $P$ inputs, $R$ outputs, $L$ hidden layers for deep CVNNs, and $N$ neurons per layer. For the deep PT-RBF, the number of bottleneck outputs is equal to the number of neurons per layer, i.e., $I^{\{l\}}=O^{\{l\}}=N$ for $l \in [1,\,2,\,\cdots,\,L-1]$, except for the output layer where $O^{\{L\}}=R$. The asymptotic computational complexities of shallow and deep CVNNs, based on Tables~\ref{tab:shallow_cvnn_comp_complexity} and \ref{tab:deep_cvnn_comp_complexity}, are depicted in Table~\ref{tab:asymptotic_cvnn_comp_complexity}. In terms of asymptotic computational complexities, both training and inference have identical results, which is why the operation mode is not addressed in Table~\ref{tab:asymptotic_cvnn_comp_complexity}. As the C-RBF and FC-RBF were only proposed for shallow architectures, their complexities are not specified for deep CVNNs.

From Table~\ref{tab:asymptotic_cvnn_comp_complexity}, for shallow CVNNs with a number of neurons much lower than the number of inputs and outputs, i.e., first column, the computational complexities are asymptotically linear. However, for shallow CVNNs with a number of neurons proportional to the number of inputs and outputs, i.e., second column, and deep CVNNs with a number of neurons per layer much higher than the number of layers, i.e., third column, the computational complexities are asymptotically quadratic. Nevertheless, the computational complexities are asymptotically cubic for deep CVNNs with a number of neurons per layer proportional to the number of layers, i.e., the fourth column.

Relying on Table~\ref{tab:asymptotic_cvnn_comp_complexity} asymptotic analysis, as shallow neural networks are usually designed with more neurons than inputs and outputs, thus shallow CVNNs have linear computational complexity when increasing the number of neurons. Notwithstanding, deep CVNNs have quadratic computational complexity with increasing the number of neurons because conventional deep neural networks operate with more neurons than hidden layers.

\begin{table*}[t]
\centering
\renewcommand{\arraystretch}{1.3}
\begin{threeparttable}
\caption{Computational Complexities of CVNNs for applications proposed in the literature.}
\label{tab:applcases_cvnn_comp_complexity}
\begin{tabular}{ccc|cc|cc|cc}
\hline
 & \multicolumn{2}{c}{\thead{MIMO channel estimation \\ and decoding \cite{Soares2021}}} & \multicolumn{2}{c}{\thead{FBMC/OQAM channel\\ estimation in IM/DD \cite{Chu2022}}} & \multicolumn{2}{c}{\thead{Beamforming receivers \\ with multiple users \cite{Mayer2022}}} & \multicolumn{2}{c}{\thead{OFDM channel estimation \\ and signal detection \cite{Ye2018}}}   \\ \cline{2-9} 
\textbf{CVNN}   & Training  & Inference & Training  & Inference     & Training  & Inference & Training  & Inference     \\
\hline
CVFNN           & 583,968   & 287,232   & 160       & 48            & 8,948     & 3,492     & 3,690,320   & 1,270,752       \\
SCFNN           & 583,904   & 287,232   & 154       & 48            & 8,942     & 3,492     & 3,690,288   & 1,270,752       \\
MLMVN           & 584,640   & 287,632   & 188       & 68            & 9,736     & 3,892     & 3,699,392   & 1,275,320       \\
C-RBF           & 429,428    & 211,300   & 187       &  65           & 4,712      & 1,900     & 700,110    & 330,038        \\
FC-RBF          & 1,268,528   & 422,400   & 432       & 120           & 12,012     & 3,600     & 1,987,144    & 657,792      \\
PT-RBF          & 449,328    & 217,800   & 312       & 100           & 54,412     & 16,400    &  7,007,408  & 2,162,668      \\ 
\hline
\end{tabular}
\end{threeparttable}
\end{table*}

\section{Use cases}
\label{sec:applcases}
To provide readers with a clear understanding, we present the computational complexities of CVNNs for some recent applications in communication systems proposed in the literature. Table~\ref{tab:applcases_cvnn_comp_complexity} depicts the computational complexities of CVNNs for MIMO channel estimation and decoding~\cite{Soares2021}, FBMC/OQAM channel estimation in intensity modulation direct detection~(IM/DD)~\cite{Chu2022}, beamforming receivers with multiple users~\cite{Mayer2022}, and OFDM channel estimation and signal detection~\cite{Ye2018}. The computational complexities of training and inference have been computed using the equations presented in Tables~\ref{tab:shallow_cvnn_comp_complexity} and \ref{tab:deep_cvnn_comp_complexity}. The CVNN architectures were determined based on the descriptions presented in each referenced work. For instance, in~\cite{Mayer2022}, the CVNNs were designed with six inputs, three outputs, and $100$~neurons. For comparison purposes, if the referenced work only discussed one CVNN architecture or exclusively employed RVNNs, we considered the number of inputs, outputs, layers, and neurons as parameters to calculate the equivalent computational complexity for the CVNNs. On the other hand, if the referenced work only considered deep architectures, we took into account the equivalent number of neurons~(i.e., the sum of all neurons) to compute the computational complexity of the shallow CVNNs, specifically C-RBF and FC-RBF.

In Table~\ref{tab:applcases_cvnn_comp_complexity}, we observe that C-RBF achieved lower computational complexities in most of the applications, with the exception of \cite{Chu2022}. When considering deep CVNN architectures, the PT-RBF presented higher computational complexity, as seen in \cite{Mayer2022, Ye2018}. On the other hand, perceptron-based CVNNs exhibit intermediate computational complexity. Based on these results, we could recommend C-RBF as the primary choice for low-power communication systems due to its lower complexity and satisfactory performance in noisy scenarios. However, in more demanding applications such as those required in base stations, the PT-RBF could also be employed, but at the cost of increased computing resources.


\section{Conclusion} 
\label{sec:conc}


This paper offers a comprehensive analysis of the computational complexities associated with various complex-valued neural network~(CVNN) architectures, including CVFNN, SCFNN, MLMVN, C-RBF, FC-RBF, and PT-RBF. Beyond simply cataloging these complexities, our work provides valuable technical insights that can guide both researchers and practitioners in the field. One of the key contributions of our analysis is the elucidation of how the asymptotic computational complexities of CVNNs evolve in relation to their architectural parameters, such as the number of inputs, outputs, neurons, and layers. By understanding these trends, practitioners can make informed decisions when selecting a CVNN architecture that aligns with their computational resource constraints. Moreover, our research goes beyond mere theoretical analysis. We demonstrate the practical utility of our findings by showcasing how quantitative computational complexities can be harnessed to accurately estimate the number of floating-point operations required for implementing CVNNs in communication systems. This insight empowers engineers and system designers to make informed choices when optimizing CVNNs for real-world applications, ultimately enhancing their efficiency and effectiveness.

\section*{Acknowledgments}

This work was supported in part by the Coordenação de Aperfeiçoamento de Pessoal de Nível Superior --- Brasil
(CAPES) --- Finance Code 001.


%

\bibliographystyle{myIEEEtran.bst}
\bibliography{references.bib}

\end{document}